\documentclass[11pt]{article} % For LaTeX2e
\usepackage{url}
\usepackage{smile}
\usepackage{graphicx} % more modern
\usepackage{epstopdf}
\usepackage{wrapfig}
\usepackage[colorlinks, linkcolor=blue, anchorcolor=blue, citecolor=blue]{hyperref}
\usepackage[margin=1in]{geometry}
\usepackage[normalem]{ulem}
\usepackage[export]{adjustbox} % http://ctan.org/pkg/adjustbox
\usepackage{mathtools, cuted}
\usepackage{enumerate}
\usepackage{enumitem}
\usepackage{microtype}
\usepackage{graphicx}
\usepackage{booktabs}
\usepackage{multirow} 
\usepackage{subcaption}
\usepackage{amsmath}
\usepackage{amssymb}
\usepackage{mathtools}
\usepackage{amsthm}
\usepackage{algorithm}
\usepackage{algorithmic}

\usepackage[margin=1in]{geometry}
\usepackage[normalem]{ulem}
\usepackage[export]{adjustbox}% http://ctan.org/pkg/adjustbox
\usepackage{mathtools, cuted}
\usepackage{natbib}
\usepackage{enumerate}
\usepackage{enumitem}

\usepackage{kpfonts}
\DeclareMathAlphabet{\mathsf}{OT1}{cmss}{m}{n}

\SetMathAlphabet{\mathsf}{bold}{OT1}{cmss}{bx}{n}

\providecommand{\norm}[1]{\|#1\|}

\newcommand{\round}{\text{round}}
\newcommand{\OurAlg}{LoftQ}

\author{Yixiao Li$^{**}$, Yifan Yu$^{**}$, Chen Liang, Pengcheng He, \\Nikos Karampatziakis, Weizhu Chen, Tuo Zhao \footnote{Li, Yu, Liang and Zhao are affiliated with Georgia Tech. He, Karampatziakisand and Chen are affiliated with Microsoft Azure. Correspondence to \url{yixiaoli@gatech.edu}, \url{yyu429@gatech.edu} and \url{tourzhao@gatech.edu}.}}

\title{LoftQ: LoRA-Fine-Tuning-Aware Quantization for Large Language Models}

\begin{document}

\maketitle

\def\thefootnote{**}\footnotetext{Equal contributions}

\begin{abstract}
% Large language models have exhibited superior performance in various language application, but they often require massive computational resources in training and deploying, especially memory. To reduce the model storage, we propose {\OurAlg} (\textbf{Lo}RA-\textbf{F}ine-\textbf{T}uning-aware \textbf{Q}uantization), a novel quantization framework that takes downstream fine-tuning into consideration and uses low-rank adaptation (LoRA) to update quantized models. 
% Our approach circumvents the intrinsic limitations of quantization, particularly its challenges related to the difficult and resource-intensive updating of quantized parameters.
% Our method alleviates the quantization errors by bridging the gap between quantized and full-precision model at initialization. 
% We evaluate our method on natural language understanding, question answering, summarization, and natural language generation tasks. Experiments show that our method is highly effective and outperforms existing quantization methods, especially in 2-bit regime. We will release our code.

% Q popular because memory low, but low bit has gap.
% This gap compromise the downstream fine-tuning, e.g., lora. 
% We propose

% We use alternating quantization + low-rank approximation to find good initialization for lora. 

Quantization is an indispensable technique for serving Large Language Models (LLMs) and has recently found its way into LoRA fine-tuning \citep{dettmers2023qlora}.
In this work we focus on the scenario where quantization and LoRA fine-tuning are applied together on a pre-trained model.  
In such cases it is common to observe a consistent gap in the performance on downstream tasks between full fine-tuning and quantization plus LoRA fine-tuning approach.
In response, we propose {\OurAlg} (\textbf{Lo}RA-\textbf{F}ine-\textbf{T}uning-aware \textbf{Q}uantization), a novel quantization framework that simultaneously quantizes an LLM and finds a proper low-rank initialization for LoRA fine-tuning. 
Such an initialization alleviates the discrepancy between the quantized and full-precision model and significantly improves generalization in downstream tasks.
We evaluate our method on natural language understanding, question answering, summarization, and natural language generation tasks. Experiments show that our method is highly effective and outperforms existing quantization methods, especially in the challenging 2-bit and 2/4-bit mixed precision regimes. 
The code is available on \url{https://github.com/yxli2123/LoftQ}.

\end{abstract}

\section{Introduction}\label{sec:introduction}
The advent of Pre-trained Language Models (PLMs) has marked a transformative shift in the field of Natural Language Processing (NLP), offering versatile solutions across various applications \citep{he2021debertav3, lewis2019bart, touvron2023llama}. They have showcased unparalleled proficiency in executing a variety of language tasks, including Natural Language Understanding (NLU) and Natural Language Generation (NLG). These models typically have millions or even billions of parameters, necessitating substantial computational and memory requirements.
% For example, LLaMA consists of up to 70 billion parameters and GPT-3 comprises up to 175 billion parameters. 
However, the extensive computational and memory demands of these models pose significant challenges, especially in real-world deployments where resources are often constrained and need to be shared among many users.
% Basically, two obstacles are in the way: the model size and the substantial associated training cost, such as gradient and optimization cache.

To mitigate the extensive storage requirements of pre-trained models, quantization serves as a pivotal compression technique \citep{zafrir2019q8bert,  shen2020q, bai2022towards, dettmers2022llm}, converting high-precision numerical values into a discrete set of values. 
Typically, model parameters, originally stored in a 16-bit float format, are transformed into a 4-bit integer format through quantization, resulting in a substantial 75\% reduction in storage overhead.
Additionally, to facilitate the adaptation of quantized pre-trained models to downstream tasks efficiently, Low-Rank Adaptation (LoRA) is a viable approach \citep{hu2021lora}. This technique is a parameter-efficient fine-tuning method traditionally applied to high-precision pre-trained models. It is based on the hypothesis that the differences between fully fine-tuned weights and pre-trained weights exhibit low-rank properties. This allows these differences to be represented using low-rank matrices. As a result, the original pre-trained weights remain unaltered, with adaptations confined solely to these low-rank matrices, enabling effective task adaptation.
% A prominent technique within the realm of quantization is Quantization-Aware Training (QAT), initially quantizes a pre-trained model and subsequently fine-tunes the quantized model on downstream tasks. 
% Even though a variety of quantization methods have been proposed to approximate the original high-precision numbers as much accurate as possible, it inevitably introduces a certain level of errors, which yield significant deviation from the original model. 

% It usually attains higher accuracy on downstream tasks compared to other quantization methods, such as post-training quantization.

% There are two major quantization approaches: Quantization-Aware Training (QAT) and Post-Training Quantization (PTQ). QAT first quantizes a pre-trained model and then fine-tunes the quantized model on downstream tasks. By contrast, PTQ calibrates a pre-trained model by a small subset of fine-tuning data and then quantizes the model based on the calibration results. 

\begin{figure}[htb!]
\centering
\begin{subfigure}{0.46\columnwidth}
    \centering
    \includegraphics[width=1.0\textwidth]{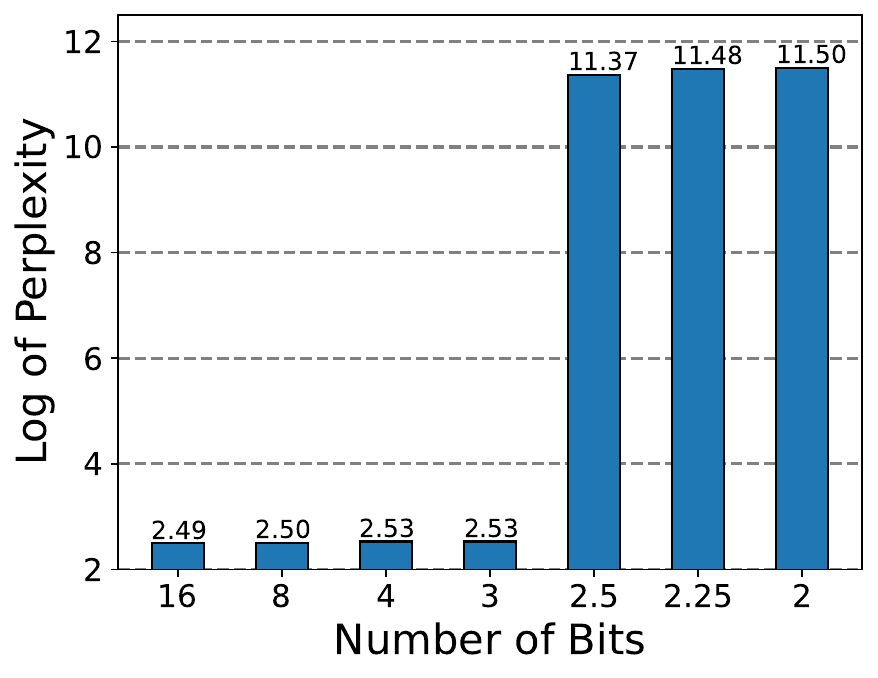}
    \vspace{-5mm}
    \caption{Pre-trained LLAMA-2-13b on WikiText-2
    \label{fig:quant_degrad_pretrain}}
\end{subfigure}
\begin{subfigure}{0.46\columnwidth}
    \centering
    \includegraphics[width=1.0\textwidth]{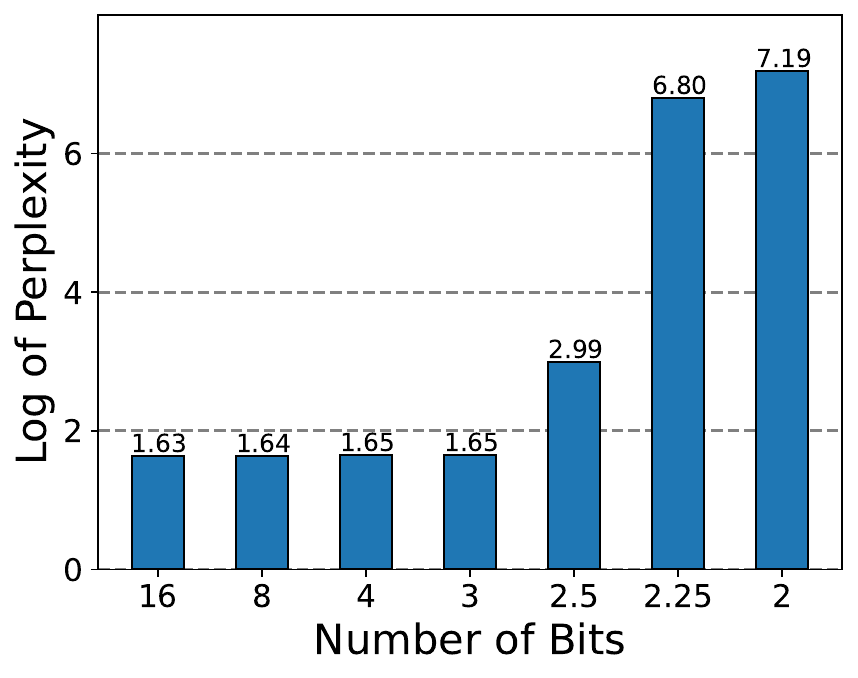}
    \vspace{-5mm}
    \caption{Fine-tuned LLAMA-2-13b on WikiText-2
    \label{fig:quant_degrad_ft}}
\end{subfigure}
\vspace{-2mm}
\caption{QLoRA performance with different bits. {\bf Left:} QLoRA initialization of LLAMA-2-13b on WikiText-2. {\bf Right:} Apply QLoRA to LLAMA-2-13b on WikiText-2 language modeling task. Smaller perplexity indicates better performance.
\vspace{-3mm}
}
\end{figure}

When quantizing pre-trained models, practitioners often concentrate primarily on the quantization technique, inadvertently neglecting the importance of subsequent LoRA fine-tuning  \citep{dettmers2023qlora, diao2023lmflow}.
For example, QLoRA inherits the fixup initialization \citep{zhang2019fixup} used in LoRA, which  \citep{dettmers2023qlora} attaches zero initialized low-rank adapters (see Section \ref{sec:background_lora}) to the quantized pre-trained model. The inevitable discrepancy introduced by quantization during the approximation of the original high-precision numbers, a scenario particularly pronounced in low-bit situations such as the 2-bit regime, can adversely impact the initialization of LoRA fine-tuning. As illustrated in Figure \ref{fig:quant_degrad_pretrain}, the quantized pre-trained model obtained by QLoRA exhibits severe degradation below the 3-bit level. This deviation in initialization often results in an inferior fine-tuning performance. As illustrated in Figure \ref{fig:quant_degrad_ft}, the fine-tuning performance drops as the quantization bit decreases when applying QLoRA. Moreover, it is noteworthy that QLoRA fails below the 3-bit level.

\begin{figure}[htb!]
\centering
\begin{subfigure}{0.48\columnwidth}
    \centering
    \includegraphics[width=1.0\textwidth]{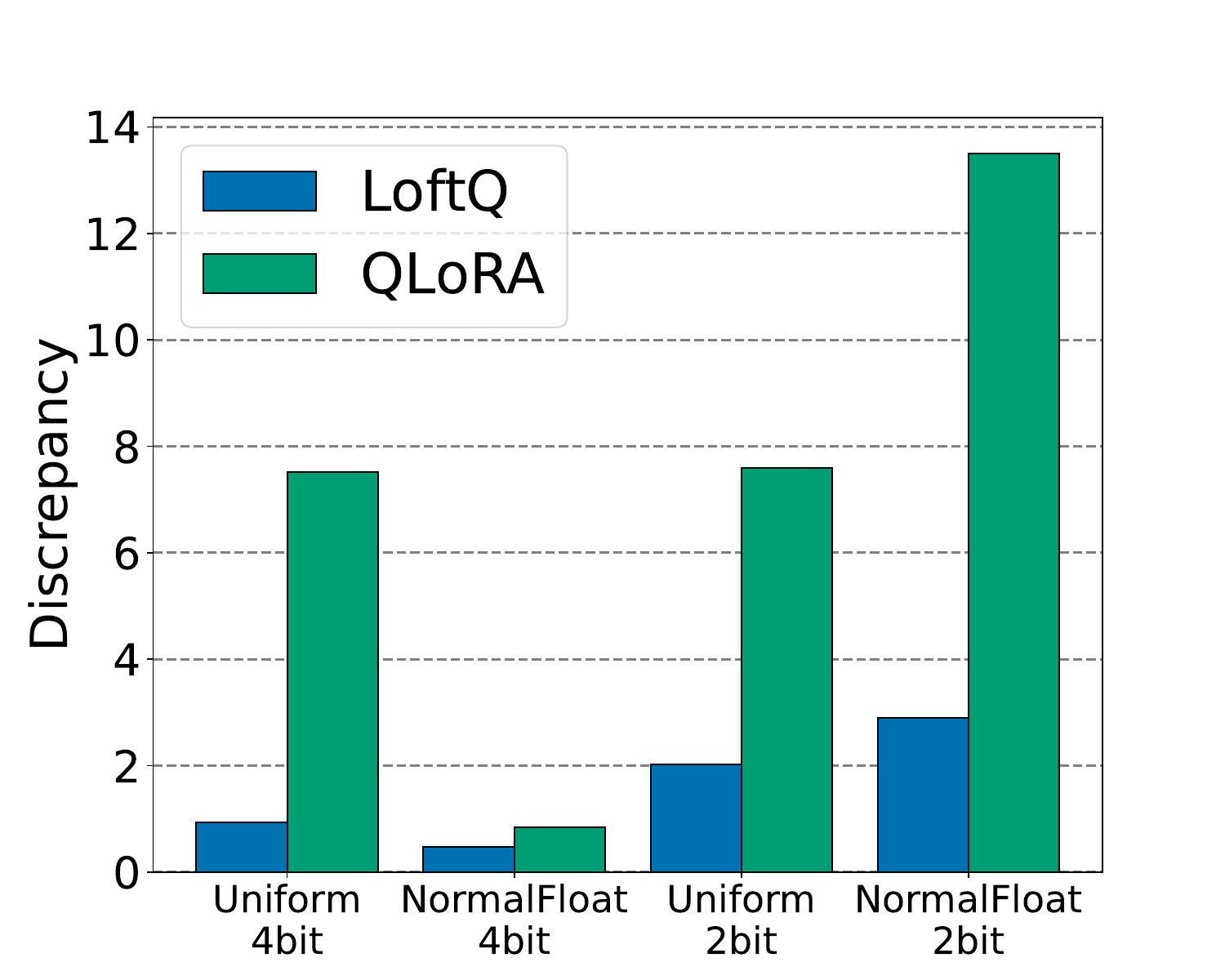}
    \vspace{-5mm}
    \caption{Spectral norm of the initialization difference}
\end{subfigure}
\begin{subfigure}{0.48\columnwidth}
    \centering
    \includegraphics[width=1.0\textwidth]{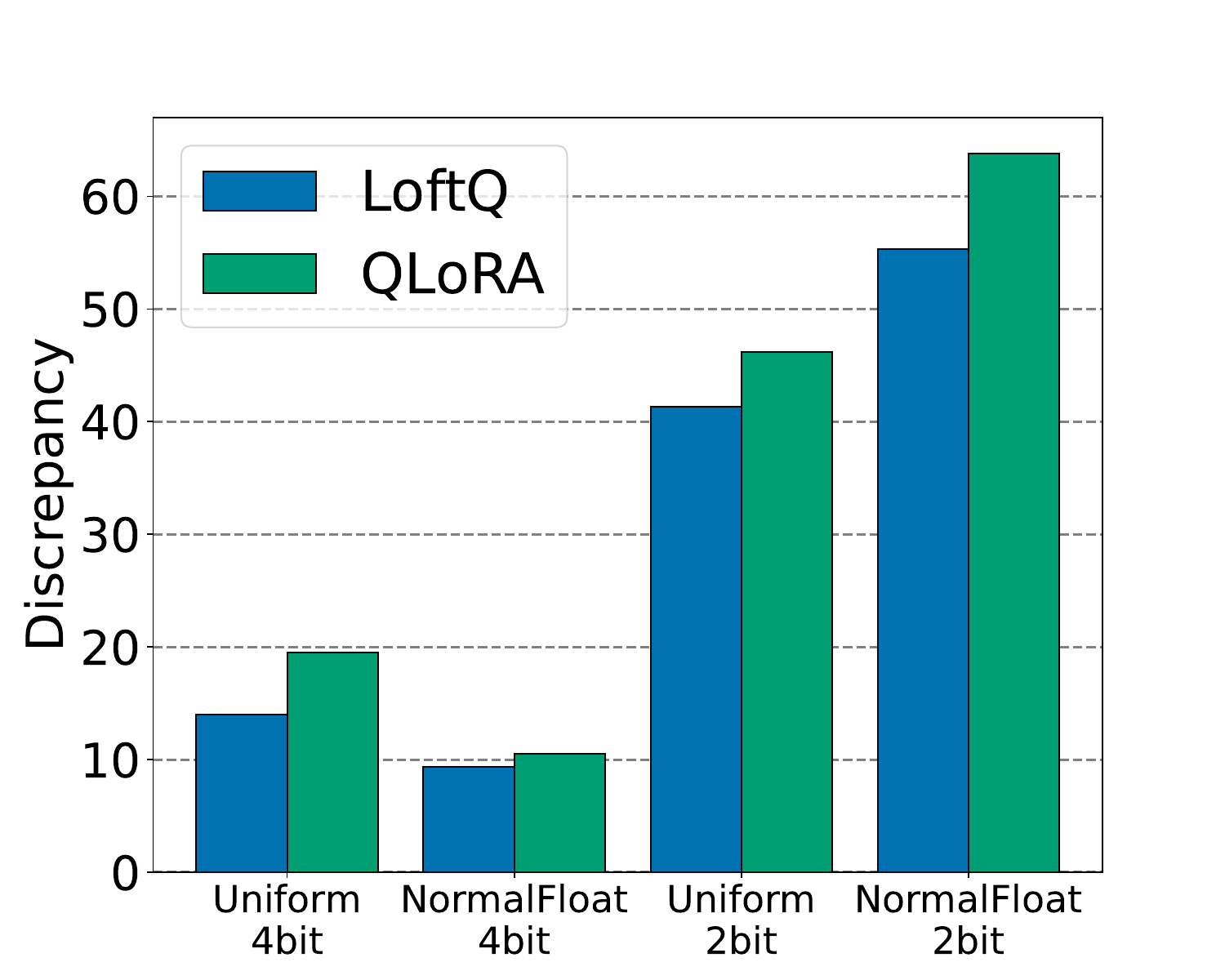}
    \vspace{-5mm}
    \caption{Frobenius norm of the initialization difference}
\end{subfigure}
\vspace{-1mm}
\caption{
Initialization discrepancy between the LoRA initialization and the original pre-trained weight matrix, described by the spectral norm and Frobenius norm of the difference. The weight matrix in the above figures is randomly selected in BART-large. The initialization is obtained by QLoRA and {\OurAlg}, with Uniform and NormalFloat quantization methods applied at both 2-bit and 4-bit levels. {\OurAlg} successfully mitigates the discrepancy, especially at the 2-bit level.
% Importance scores are calculated when compressing DeBERTa-base on SST-2 at the 3156\textsuperscript{th} step. \textit{Layer 7, Query} indicates the matrix is from the query matrix at the 7th layer.
}
\vspace{-1mm}
\label{fig:quant_error}
\end{figure}

In this paper, we introduce a novel quantization framework, called \textbf{Lo}RA-\textbf{F}ine-\textbf{T}uning-aware \textbf{Q}uantization (LoftQ). It is designed specifically for pre-trained models that require quantization and LoRA fine-tuning.
This framework actively integrates low-rank approximation, working in tandem with quantization to jointly approximate the original high-precision pre-trained weights. This synergy significantly enhances alignment with the original pre-trained weights as illustrated in Figure~\ref{fig:quant_error}. 
Consequently, our method provides an advantageous initialization point for subsequent LoRA fine-tuning, leading to improvements in downstream tasks.

We evaluate our quantization framework by conducting extensive experiments on downstream tasks, such as NLU, question answering, summarization, and NLG. 
Experiments show that {\OurAlg} consistently outperforms QLoRA across all precision levels. For instance, with 4-bit quantization, we achieve a 1.1 and 0.8 gain in Rouge-1 for XSum \citep{Narayan2018xsum} and CNN/DailyMail \citep{hermann2015teaching}, respectively. {\OurAlg} excels particularly in low-bit scenarios and works effectively with different quantization methods. For example, we achieve over an 8\% gain on MNLI \citep{wang2018glue} and more than 10\% on SQuADv1.1 \citep{rajpurkar-etal-2016-squad} with both 2-bit NormalFloat and the 2-bit uniform quantization. We have not seen our approach performs worse than QLoRA.

% \textbf{Pruning} is also widely used to reduce the model size. It removes redundant parameters based on the importance score, for example, magnitude \citep{han2015learning}, sensitivity \citep{sanh2020movement, molchanov2019importance}, and uncertainty \citep{zhang2022platon}. However, when subjected to extreme high compression rates, such as removing more than 80\% parameters, the performance will demonstrate notable degradation. 
% We will compare our quantization method with the state-of-the-art pruning method \citep{li2023losparse} in Section \ref{sec:experiment}.\red{(need to mentioned it?)}

\section{Background}
\subsection{Transformer Models}
A transformer model contains a sequence of layers, where each layer consists of two sub-layers: a multi-head self-attention (MHA) and a fully connected feed forward network (FFN) \citep{vaswani2017attention}. 
 % Given an input $X \in \RR^{d_1}$, the output $Y \in \RR^{d_2}$ for each layer is computed as
Given the input $X \in \RR^{n\times d}$, where $n$ is the sequence length and $d$ is the hidden dimension of the model, MHA computes the $h$ attention heads in parallel: 
\begin{align*}
    & \mathrm{MHA}(X) = \mathrm{Concat}(\mathrm{head}_1, ..., \mathrm{head}_h) W_o,\\
    \text{where}~~~ \mathrm{head}_i = &\mathrm{Softmax}({XW_{q_i} (XW_{k_i})^\top}/{\sqrt{d_h}}) XW_{v_i} ~~\text{for}~~~~i = 1, ..., h,
\end{align*}
where $W_{q_i}, W_{k_i}, W_{v_i} \in \RR^{d \times d_h}$ are query, key, and value matrices, $W_o \in \RR^{d\times d}$ is the output matrix, and $d_h = d/h$. FFN comprises two linear transformations and an activation function, and is defined as $ \mathrm{FFN}(X) = \sigma(X W_{f_1} + b_1) W_{f_2} + b_2, $
where $W_{f_1} \in \RR^{d \times d_m} $, $ W_{f_2} \in \RR^{d_m \times d}$, and $\sigma(\cdot)$ is the activation function. A residual connection is used and followed by layer normalization.     

\subsection{Quantization}
{\bf Quantization.} Given a high-precision number, e.g., such as 32-bit floating point number, $X^{\text{HP}} \in \RR$, $N$-bit quantization encodes it to an integer $X^{\text{INT}} \in \{0, 1, ..., 2^{N}-1\}$. This process can be expressed as
\begin{align}
\label{eq:quant}
    X^{\text{INT}} = \round\left((2^{N} - 1) F\left( X^{\text{HP}} \right) \right),
\end{align}
where $F(\cdot) \colon \RR \mapsto [0, 1]$ is a normalization function. Uniform quantization assumes $F(X) = (X - X_{\min}) /  (X_{\max} - X_{\min})$. \citet{dettmers2023qlora} proposes 4-bit NormalFloat Quantization (NF4). It assumes $X \sim \cN(0, \sigma^2)$ and hence $F(X) = \Phi(X/\sigma)$, where $\Phi(\cdot)$ is the cumulative distribution function of the standard normal distribution. 

\noindent {\bf Dequantization.} A lookup table $\cT$, where
\begin{align}
\label{eq:lookup_table}
    \cT[i] = F^{-1}\left( \frac{i}{2^N - 1} \right), i = 0, 1, ..., 2^{N}-1,
\end{align}
is used to decode the integer $X^{\text{INT}}$ to its simulated high-precision counterpart $X^{\text{D}} \in \RR$.
Therefore, the dequantization can be expressed as 
\begin{align}
\label{eq:dequant}
    X^{\text{D}} = \cT[X^{\text{INT}}].
\end{align}

\noindent {\bf Simulated Quantization for Matrices.} While it is possible to perform multiplication directly between quantized representations, it is common to apply simulated quantization for matrices \citep{bai2020binarybert, shen2020q}.
There, quantized weight matrices are stored as encoded integers in memory, and are temporarily dequantized to simulated high-precision matrices by the lookup table when engaged in multiplication operations.
In simulated quantization, it is only necessary to analyze the map from a high-precision matrix to a simulated high-precision matrix. We denote this end-to-end process by $q_N(\cdot) \colon \RR^{m \times n} \mapsto \RR_{N}^{m \times n}$, where $\RR_{N}: \{\cT[i] \in \RR | 0 \leq i < 2^N \}$. 
% that maps a high-precision matrix $W^{\text{HP}}$ to a simulated high-precision matrix $W^{\text{D}}$. 

% While it is possible to perform multiplication directly between quantized representations, we focus on simulated quantization.
% There, quantized numbers are stored using a low-precision format in memory, and are dequantized to a high-precision format only when engaged in multiplication operations. 
% Specifically, dequantization maps $X^{\text{LP}}$, an integer, back to a simulated value $X^{\text{D}} \in \RR$ to approximate $X^{\text{HP}}$ by
% \begin{align}
%     X^{\text{D}} &= F^{-1}\left( \frac{X^{\text{LP}}}{2^N - 1} \right) \\
%     &= a[X^{\text{LP}}],
% \end{align}
% where $a \in \RR^{2^N}$ is a lookup table that stores the high-precision counterparts of low-precision values, and $X^{\text{D}}$ is the $X^{\text{LP}}$-th component in $a$.

\subsection{Low-Rank Adaptation}
\label{sec:background_lora}
LoRA \citep{hu2021lora} updates two small weight matrices $A$ and $B$ that are attached to a frozen pre-trained weight matrix $W$. Hence, a linear transformation, $Y = XW$, is reformulated as
\begin{align}
\label{eq:lora}
    Y = XW + XAB^{\top},
\end{align}
where $X \in \RR^{n \times d_1}, W \in \RR^{d_1 \times d_2}, A \in \RR^{d_1 \times r}, B \in \RR^{d_2 \times r}$, and $r \ll \min\{d_1, d_2\}$. Initially, 
\begin{align}
\label{eq:lora_init}
    A \sim \cN (0, \sigma^2), ~ B = 0,
\end{align}
so as to align to the pre-trained weights. During the fine-tuning, $W$ is fixed while $A$ and $B$ are updated by some SGD-type optimization method.

It is worth noting that if low-rank adapters $A$ and $B$ are attached to a quantized backbone $Q=q_N(W)$ and are initialized by \eqref{eq:lora_init}, the starting weight $Q + AB^{\top}$ is no longer equal to the pre-trained weight $W$ due to the discrepancy introduced by the quantization. 
% \citet{Liao2023MakeYP} have shown that too much deviation from the original pre-trained weights causes performance degradation on downstream fine-tuning tasks.

\section{Method}\label{sec:method}
We propose \textbf{Lo}RA-\textbf{F}ine-\textbf{T}uning-aware \textbf{Q}uantization (LoftQ), a quantization framework for LLMs. It alternatively applies quantization and low-rank approximation to approximate original pre-trained weights. 
This quantization framework provides a promising initialization for LoRA fine-tuning, which alleviates the quantization discrepancy in QLoRA and improves generalization in downstream tasks significantly.

\subsection{LoRA-Aware Quantization}
% Given an $N$-bit low-precision weight $Q \in \RR_{N}^{d_1 \times d_2}$ that is quantized from a high-precision weight $W \in \RR^{d_1 \times d_2}$, we add low-rank adapters $A \in \RR^{d_1 \times r}$ and $B \in \RR^{d_2 \times r}$ to it. 
% Usually, these low-rank adapters are initialized by \ref{eq:lora_init}. Such an initialization, $Q + AB^{\top}$, however, no longer aligns with the high-precision pre-trained weights due to the quantization error. This misalignment often leads to severe performance decrease during fine-tuning, especially in low-bit situation (see Section \ref{sec:experiment}).
We use an $N$-bit quantized weight $Q \in \RR_{N}^{d_1 \times d_2}$ and low-rank approximations $A \in \RR^{d_1 \times r}, B \in \RR^{d_2 \times r}$ to approximate the original high-precision pre-trained weight $W \in \RR^{d_1 \times d_2}$ as the initialization of LoRA fine-tuning.
Specifically, before fine-tuning, we initialize the network by minimizing the following objective:
\begin{align}
\label{eq:optimization}
   \underset{Q, A, B}{\min}  \norm{W - Q - AB^{\top}}_{F},
\end{align}
where $\norm{\cdot}_{F}$ denotes the Frobenious norm. This objective in \eqref{eq:optimization} takes LoRA fine-tuning into consideration by jointly optimizing the initial values of the quantized backbone $Q$ and low-rank adapters $A, B$.
% The optimal $Q$ is the frozen backbone and the optimal $A, B$ servers as the low-rank adapters in LoRA fine-tuning. 
Contrarily, practitioners typically convert the pre-trained weight $W$ into a quantized weight $Q$ outright, neglecting the subsequent LoRA fine-tuning process. This oversight leads to notable performance degradation in downstream tasks arising from the quantization discrepancy.
%This lack of consideration results in a substantial performance drop on downstream tasks due to the quantization error.

% Taking LoRA fine-tuning into consideration, we introduce low-rank approximation  $A \in \RR^{d_1 \times r}$ and $B \in \RR^{d_2 \times r}$. During LoRA initialization, our goal is to consider the quantization and low-rank approximation as a unified entity and align it with the original high-precision weights. 

% We solve the optimization in \eqref{eq:optimization} by combining quantization and singular value decomposition (SVD).

\subsection{Alternating Optimization}
\label{sec:alternating_optimization}
We solve the minimization problem in \eqref{eq:optimization} by alternating between quantization and singular value decomposition (SVD). To begin with, we set $A_0$, and $B_0$ equal to 0.

\noindent \textbf{Quantization}. At the $t$-th step, we quantize the difference between the original pre-trained weight $W$ and the low-rank approximation $A_{t-1}B_{t-1}^{\top}$ from the last step to obtain the quantized weight $Q_{t}$ by 
\begin{align}
\label{eq:quant_method}
    Q_t = q_N(W - A_{t-1}B_{t-1}^{\top}),
\end{align}
where $q_N(\cdot)$ maps a high-precision weight matrix to a quantized matrix.

We remark that our algorithm is compatible with different quantization functions $q_N(\cdot)$. We apply NF4 and the uniform quantization in Section \ref{sec:experiment} as examples. We also remark that $Q_t$ is not an exact solution of the minimization in \eqref{eq:optimization}, given the fixed $A_{t-1}B_{t-1}^{\top}$, but it is an efficient approximation.

\noindent \textbf{SVD}. After obtaining the $t$-th quantized weight $Q_t$, SVD is applied to the residual of the quantization denoted by $R_t = W - Q_t$ by
\begin{align}\label{eq:residual}
R_t =\sum_{i=1}^d\sigma_{t, i} u_{t, i} v_{t,i}^{\top},
\end{align}
where $d = \min\{d_1,d_2\}$, $\sigma_{t, 1} \geq \sigma_{t, 2} \geq ... \geq \sigma_{t, d}$ are the singular values of $R_t$, $u_{t, i}$'s and $v_{t, i}$'s are the associated left and right singular vectors of $R_t$. We then obtain a rank-$r$ approximation of $R_t$ by $A_{t}B_{t}^{\top}$, where
\begin{align}
\label{eq:svd}
A_t &= [\sqrt{\sigma_{t, 1}} u_{t, 1},...,\sqrt{\sigma_{t, r}}u_{t, r}],\nonumber\\
B_t &= [\sqrt{\sigma_{t, 1}} v_{t, 1},...,\sqrt{\sigma_{t, r}}v_{t, r}].
\end{align}

We summarize our method in Algorithm \ref{alg:alter}. It is worth noting that $T=1$ is a special case where $Q_1$ is the exact quantized weight obtained by QLoRA, and low-rank approximations $A_1, B_1$ are obtained by the SVD of the quantization residual $W - Q_1$. $T=1$ is sufficient to mitigate the quantization discrepancy, and alternating optimization helps to find a closer initialization to the pre-trained weight $W$, which further improves the performance (see Section \ref{fig:num_iter}).

\begin{algorithm}
    \caption{{\OurAlg}}\label{alg:alter}
    \begin{algorithmic}[1]
        \INPUT {Pre-trained weight $W$, target rank $r$, $N$-bit quantization function $q_N(\cdot)$, alternating step $T$}
        \STATE {Initialize $A_0 \leftarrow 0, B_0 \leftarrow 0$}
        \FOR {t = $1$ to $T$}
            \STATE {Obtain quantized weight $Q_t \leftarrow q_N(W - A_{t-1}B_{t-1}^{\top})$}
            \STATE {Obtain low-rank approximation $A_t, B_t \leftarrow \text{SVD}(W - Q_t)$} by \eqref{eq:svd}{}
            
        \ENDFOR
        \OUTPUT {$Q_T, A_T, B_T$}
    \end{algorithmic}
\end{algorithm}

% Figure \ref{fig:optimization_singular_values} shows that the residual $R$ in \eqref{eq:residual} has only a few large singular values, demonstrating its low-rank characteristics and indicating the low-rank approximation is efficient.

% We remark that our method is applicable to different types of quantization methods such as NF4 and uniform. 
% the quantization function $q_N(\cdot)$ in \eqref{eq:quant_method} is not limited to some specific ones, such as NF4, the one used in QLoRA. Instead, we can apply various quantization method, for example, uniform quantization. The experiments in Section \ref{sec:experiment} show the success of our method in a variety of quantization methods. 

We remark that the computational cost of {\OurAlg} is negligible because it is applied to individual weight matrices and therefore can be executed in parallel. We also remark one can apply {\OurAlg} only once to a pre-trained model and reuse the initialization obtained by {\OurAlg} for different downstream tasks.
% For example, it takes less than an hour to apply {\OurAlg} to LLAMA-2-13b on CPU.

% From Figure \ref{fig:alter_loss}, we notice the alternating method successfully further reduce the discrepancy between the pre-trained weights and the initialization. 

% \begin{figure}[htb!]
% \centering
% \begin{subfigure}{0.49\columnwidth}
%     \centering
%     \includegraphics[width=1.0\textwidth]{figures/alter_loss_bart.pdf}
%     \caption{$W_{\rm{fc1}}$ of Layer 0 of BART-large}
%     \label{fig:alter_loss}
% \end{subfigure}
% \begin{subfigure}{0.49\columnwidth}
%     \centering
%     \includegraphics[width=1.0\textwidth]{figures/bart-decoder-fc2-nf2-lora.pdf}
%     \caption{$W_{\rm{fc2}}$ of Layer 2 of BART-large}
%     \label{fig:optimization_singular_values}
% \end{subfigure}
% \caption{
% Illustration of the effectiveness of alternative optimization. \textbf{Left:} Alternative optimization successfully reduces the spectral norm of the difference between pre-trained weights and hybrid approximation. \textbf{Right:} The residual only displays few large eigenvalues where the number of large eigenvalues equals our designated rank. This low rank approximation for the residual is efficient.
% % Importance scores are calculated when compressing DeBERTa-base on SST-2 at the 3156\textsuperscript{th} step. \textit{Layer 7, Query} indicates the matrix is from the query matrix at the 7th layer.
% }
% \vspace{-1mm}
% \label{fig:optimization}
% \end{figure}

\subsection{Applying to LoRA Fine-tuning}
We store the $Q_T \in \RR_N^{d_1 \times d_2}$ obtained by {\OurAlg} using an integer matrix $M$ by \eqref{eq:quant} and a lookup table $\cT$ by \eqref{eq:lookup_table}. 
% The integer matrix $M$ serves as the frozen backbone weight in LoRA fine-tuning. We initialize the low-rank adapter of LoRA with $A_T, B_T$ obtained by {\OurAlg}.
We initialize the backbone with the integer matrix $M$ and initialize the low-rank adapters with $A_T, B_T$ obtained by {\OurAlg}. 

During LoRA fine-tuning, we freeze the integer weight $M$ and optimize the low-rank adapters with an efficient optimization algorithm, e.g., AdamW \citep{loshchilov2017decoupled}.
% We apply $Q_T$ and $A_T, B_T$ obtained by {\OurAlg} as the initialization of LoRA fine-tuning. $Q_T$ serves as the frozen backbone weight, and $A_T, B_T$ are the initial low-rank adapter.
% In LoRA fine-tuning, we fix the quantized weights $Q_T$ and update low-rank adapters starting from $A_T, B_T$ by some SGD-type optimization algorithm.
% In LoRA fine-tuning, we fix the integer weight $M$ and update low-rank adapters $A, B$ starting from $A_T, B_T$ by some SGD-type optimization algorithm.
% In forward propagation, the integer weight $M$ is temporarily dequantized to a simulated high-precision weight by its lookup table when it is engaged in matrix multiplication.
In forward propagation, the integer weight $M$ is temporarily dequantized to the simulated high-precision weight $Q_T$ by its lookup table, as described in \eqref{eq:dequant}.
% Such temporary dequantization does not cost additional storage when the linear transformation has completed.
% rendered to high precision just in time, or use special low bit matrix multiplication code such as \texttt{bitsandbytes}~\cite{dettmers2022llmint8,dettmers2022optimizers}.
In back propagation, gradients and optimizer state are only related to low-rank adapters $A, B$, which reduces considerable training cost. 

\section{Experiments}
\label{sec:experiment}
We evaluate our method on NLU and NLG tasks. We apply {\OurAlg} for quantizing DeBERTaV3-base \citep{he2021debertav3}, BART-large \citep{lewis2019bart}, and LLAMA-2 series \citep{touvron2023llama}.

\noindent \textbf{Implementation Details.} Following the prior works of LoRA variants \citep{zhang2023adaptive, he2021towards}, we freeze all the backbone weight matrices and add low-rank adapters to weight matrices in MHA and FFN of all layers. We quantize the weight matrices that are attached by low-rank adapters. All the quantized models and adapters used in this paper are available on \url{https://huggingface.co/LoftQ}.
% Given that NLU is a relatively easy task, %we primarily focus on low-bit regime. 
% we also apply quantization and low-rank adapters to the embedding layers to further enhance the compression efficiency. %when conducting NLU tasks. 
% For more challenging tasks in NLG, we freeze the embedding layers without applying quantization or low-rank adapters.
Our implementation is based on publicly available {\it Huggingface Transformers} code-base \citep{NEURIPS2019_9015}. All the experiments are conducted on NVIDIA A100 GPUs.
% \footnote{We don't quantize the embedding of BART-large and LLAMA-2 in our experiments. However, we find quantized embedding doesn't harm the performance of DeBERTaV3-base on downstream tasks. Therefore, we quantize the embedding of DeBERTaV3-base on natural language understanding tasks.}

\noindent \textbf{Quantization Methods.} We apply two quantization methods to demonstrate {\OurAlg} is compatible with different quantization functions:
\begin{itemize}
    \item \textit{Uniform quantization} is a classic quantization method. It uniformly divides a continuous interval into $2^N$ categories and stores a local maximum absolute value for dequantization.
    \item \textit{NF4} and its 2-bit variant \textit{NF2} are quantization methods used in QLoRA \citep{dettmers2023qlora}. They assume that the high-precision values are drawn from a Gaussian distribution and map these values to discrete slots that have equal probability.
\end{itemize}
%\vspace{-2mm}
We perform 2-bit and 4-bit quantization on all models, achieving compression ratios of 25-30\% and 15-20\% at the 4-bit and 2-bit levels, respectively. The compression ratios and trainable parameter ratios for all models are detailed in the Appendix \ref{app:model_compression_ratio}.

\noindent \textbf{Baselines.} We compare {\OurAlg} with the following baseline methods: 
\begin{itemize}
    \item \textit{Full fine-tuning} is the most common approach for adapting a pre-trained model to downstream tasks. The model is initialized with pre-trained weights and all parameters are updated through an SGD-type optimization method. 
    \item \textit{Full precision LoRA (LoRA)} is a lightweight method for task adaptation, where it stores the backbone using 16-bit numbers and optimizes the low-rank adaptors only. The adaptors are applied to the same matrices as in {\OurAlg}. %, where its backbone is stored using 16-bit numbers, is a light method to adapt pre-trained models to downstream tasks. 
    % It is applied to the same matrices as {\OurAlg}.
    \item \textit{QLoRA} is similar to \textit{LoRA} except the backbone is quantized into low-bit regime. The low-rank adapters are initialized using \eqref{eq:lora_init} and are applied to the same matrices as in {\OurAlg}.% It attaches the low-rank adapters that are initialized by \eqref{eq:lora_init} to the same matrices as {\OurAlg}. To achieve a higher efficiency, we also apply quantization and low-rank adapters to the embedding layers in NLU tasks.
    % \item \textit{QTA} (?*) .
\end{itemize}

% \textbf{Models.} We present both compression ratios and trainable parameter ratios in Appendix \ref{app:model_compression_ratio}. We achieve 25-30\% compression ratios for 4-bits quantization, and we achieve 15-20\% compression ratios at 2-bits level. 

% Our implementation is based on publicly available {\it Huggingface Transformers} code-base \citep{NEURIPS2019_9015}. All the experiments are conducted on NVIDIA A100 GPUs. 

\subsection{Encoder-only Model: DeBERTaV3}

\textbf{Models and Datasets.}
We quantize the DeBERTaV3-base \citep{he2021debertav3} with {\OurAlg}, then finetune and evaluate the model on the General Language Understanding Evaluation (GLUE) benchmark \citep{wang2018glue}, SQuADv1.1 \citep{rajpurkar-etal-2016-squad}, and ANLI \citep{Nie2019AdversarialNA}. The specific tasks of GLUE are given in Appendix \ref{app:dataset-glue}. Following previous works \citep{zhang2023adaptive}, we exclude WNLI in the experiments.

\noindent \textbf{Implementation Details.}
We select the learning rates from $\{1\times10^{-5},5\times10^{-5} , 1\times10^{-4}\,5\times10^{-4}\}$. We quantize the entire backbone. Given that GLUE,  SQuADv1.1, and ANLI are relatively easy NLU tasks, we also quantize the embedding layer for higher compression efficiency. We apply the NormalFloat and the uniform quantization for {\OurAlg} and QLoRA at both 2-bit and 4-bit levels. We use rank 16 and 32 for low-rank adapters. %We don't present extensive experiments using 4-bit quantization on GLUE, as QLoRA with NF4 quantization has already demonstrated performance comparable to full fine-tuning on GLUE. %We focus our 4-bit experiments on more challenging tasks, specifically summarization and generation tasks, utilizing models such as BART-large and LLAMA-2.
%}. We conduct experiments with both uniform quantization and NF2 quantization. 
More implementation details, such as the training epochs and batch sizes, are presented in Appendix \ref{sec:app_nlu_details}.

\noindent \textbf{Main Results.} Table \ref{tab:glue_datasets_normal_float} and Table \ref{tab:glue_datasets_uniform} summarize the results for 2-bit quantization on the GLUE, SQuADv1.1, and ANLI datasets, by NF2 and the uniform quantization, respectively.
Our method consistently outperforms QLoRA on all settings with respect to different ranks, quantization methods, and datasets. When using the uniform quantization (Table \ref{tab:glue_datasets_uniform}), our method achieves 88.0\% accuracy on MNLI-m, surpassing the QLoRA baseline by 8\%. For tasks like SST and SQuADv1.1, our method even approaches the full fine-tuning performance at 2-bit level. The 4-bit quantization experiment results are presented in Appendix~\ref{sec: app_nlu_nf4} as both {\OurAlg} and QLoRA achieve performance close to full fine-tuning.
 % We remark though quantization with normal float quantization using 4-bits achieves superior performance that is close to full-fintuning\citep{dettmers2023qlora}, normal float quantization using 2-bits performs worse than uniform quantization under 2-bits settings. This indicates different precisions require different quantization methods.
\begin{table*}[thb!]
%\vspace{-4mm}
\caption{Results with 2-bit {\OurAlg} of DeBERTaV3-base models on GLUE development set, SQuADv1.1 development set, ANLI test set using \text{\bf NF2 quantization}. We report the median over four seeds. {\it N.A.} indicates the model does not converge. The best results on each dataset are shown in {\bf bold}.}
\vspace{-4mm}
\label{tab:glue_datasets_normal_float}
\begin{center}
\scalebox{0.82}{
\begin{small}
\begin{tabular}{c|c|cccccccccc}
\toprule
{\bf Rank} & {\bf Method} & {\bf MNLI} & {\bf QNLI} & {\bf RTE} & {\bf SST} & {\bf MRPC} & {\bf CoLA} & {\bf QQP} & {\bf STSB}&{\bf SQuAD} &{\bf ANLI}\\ 
 ~ & ~ & {m / mm} & {Acc} & {Acc} & {Acc} & {Acc} & {Matt} & {Acc} & { P/S Corr}&{EM/F1} &{Acc}\\
\midrule 
 
- & {Full FT} & {90.5/90.6} & {94.0} & {82.0} & {95.3} & {89.5/93.3} & {69.2} & {92.4/89.8} & {91.6/91.1} & {88.5/92.8} & 59.8\\ \midrule
 {16} & {LoRA} &  {90.4/90.5} & {94.6} & {85.1} & {95.1} & {89.9/93.6} & {69.9} & {92.0/89.4} & {91.7/91.1} & {87.3/93.1} & 60.2\\
\midrule
 \multirow{2}*{16} & {QLoRA} & {75.4/75.6} & {82.4}&{55.9}  & {86.5} & {73.8/82.8}&{N.A.}&{86.8/82.3}&{83.0/82.8} & {61.5 / 71.2}&{N.A.} \\
 ~ & {\OurAlg} &{\bf 84.7/85.1} & {\bf86.6}&{\bf 61.4} & {\bf90.2} & {\bf83.8/88.6}&{\bf37.4}&{\bf90.3/86.9}&{\bf87.1/86.9} & {\bf81.5/88.6} & {\bf47.1} \\ \cmidrule{1-12}
 \multirow{2}*{32} & {QLoRA} & {78.5/78.7} & {80.4} & {56.7} & {86.9} & {73.8/82.7} & {N.A.}&{87.1/82.7} & {83.6/83.3} & {64.6/73.8}& {N.A.} \\
 ~ & {\OurAlg} & {\bf 86.0/86.1}&{\bf 89.9} & {\bf 61.7} & {\bf 92.0} & {\bf 83.6/87.2} & {\bf 47.5}&{\bf 91.0/87.9} & {\bf 87.5/87.0} & {\bf 82.9/89.8}&{\bf 49.0} \\
% \midrule
% \multirow{4}*{Uniform} & \multirow{2}*{16} & {QLoRA} & {76.5/76.3} & {83.8} &{56.7}& {86.6} & {75.7/84.7}&{\emph{N.A.}} & {87.1/82.6} & {83.5/83.4} \\
% ~ & ~ & {\OurAlg}& {\bf 87.3/87.1} & {\bf 90.6} &{\bf 61.1}& {\bf 94.0} & {\bf 87.0/90.6}&{\bf 59.1} & {\bf 90.9/88.0} & {\bf 87.9/87.6} \\ \cmidrule{2-11}
% ~ & \multirow{2}*{32} & {QLoRA} & {79.9/79.5} & {83.7} & {57.8} & {86.9} & {76.5/84.5} & {\emph{N.A.}} & {88.6/84.7} & {84.1/84.0}\\
% ~ & ~ & {\OurAlg} & {\bf88.0/88.1} & {\bf92.4} & {\bf63.2} & {\bf94.7} & {\bf87.5/91.2} & {\bf60.5} & {\bf91.3/88.3} & {\bf89.5/89.2} \\
\bottomrule
\end{tabular}
\end{small}}
\end{center}
\vspace{-2mm}
\end{table*}

\begin{table*}[thb!]
%\vspace{-1mm}
\caption{Results with 2-bit {\OurAlg} of DeBERTaV3-base models on GLUE development set, SQuADv1.1 development set using \textbf{Uniform quantization} . We report the median over four seeds. {\it N.A.} indicates the model does not converge. The best results on each task are shown in {\bf bold}.}
\vspace{-4mm}
\label{tab:glue_datasets_uniform}
\begin{center}
\scalebox{0.87}{
\begin{small}
\begin{tabular}{c|c|ccccccccc}
\toprule
 {\bf Rank} & {\bf Method} & {\bf MNLI} & {\bf QNLI} & {\bf RTE} & {\bf SST} & {\bf MRPC} & {\bf CoLA} & {\bf QQP} & {\bf STSB} &{\bf SQuAD} \\ 
 ~ & ~ & {m / mm} & {Acc} & {Acc} & {Acc} & {Acc} & {Matt} & {Acc} & { P/S Corr} & {Em/F1}\\
\midrule 

- & {Full FT} & {90.5/90.6} & {94.0} & {82.0} & {95.3} & {89.5/93.3} & {69.2} & {92.4/89.8} & {91.6/91.1} & {88.5/92.8}\\ \midrule
 {16} & {LoRA} &  {90.4/90.5} & {94.6} & {85.1} & {95.1} & {89.9/93.6} & {69.9} & {92.0/89.4} & {91.7/91.1} & {87.3/93.1}\\
\midrule
% \multirow{4}*{NF4} & \multirow{2}*{16} & {QLoRA} & {75.4/75.6} & {82.4}&{55.9}  & {86.5} & {73.8 / 82.8}&{\emph{N.A.}}&{86.8 / 82.3}&{83.0/82.8}\\
% ~ & ~ & {\OurAlg} &{\bf 84.7/85.1} & {\bf86.6}&{\bf 61.4} & {\bf90.2} & {\bf83.8 /88.6}&{\bf37.4}&{\bf90.3 / 86.9}&{\bf87.1 / 86.9} \\ \cmidrule{2-11}
% ~ & \multirow{2}*{32} & {QLoRA} & {78.5/78.7} & {80.4} & {56.7} & {86.9} & {73.8/82.7} & {\emph{N.A.}}&{87.1/82.7} & {83.6/83.3} \\
% ~ & ~ & {\OurAlg} & {\bf 86.0/86.1}&{\bf 89.9} & {\bf 61.7} & {\bf 92.0} & {\bf 83.6/87.2} & {\bf 47.5}&{\bf 91.0/87.9} & {\bf 87.5/87.0}  \\
% \midrule
\multirow{2}*{16} & {QLoRA} & {76.5/76.3} & {83.8} &{56.7}& {86.6} & {75.7/84.7}&{N.A.} & {87.1/82.6} & {83.5/83.4}&{69.5/77.6} \\
 ~ & {\OurAlg}& {\bf 87.3/87.1} & {\bf 90.6} &{\bf 61.1}& {\bf 94.0} & {\bf 87.0/90.6}&{\bf 59.1} & {\bf 90.9/88.0} & {\bf 87.9/87.6}&{\bf84.4/91.2} \\ \cmidrule{1-11}
 \multirow{2}*{32} & {QLoRA} & {79.9/79.5} & {83.7} & {57.8} & {86.9} & {76.5/84.5} & {N.A.} & {88.6/84.7} & {84.1/84.0}&{71.6/80.2}\\
 ~ & {\OurAlg} & {\bf88.0/88.1} & {\bf92.2} & {\bf63.2} & {\bf94.7} & {\bf87.5/91.2} & {\bf60.5} & {\bf91.3/88.3} & {\bf89.5/89.2}&{\bf85.2/91.6} \\
\bottomrule
\end{tabular}
\end{small}}
\end{center}
\vspace{-2mm}
\end{table*}

Our method is also more stable compared to QLoRA in the low-bit regime. For instance, while QLoRA fails to converge on CoLA for both quantization methods and ranks, {\OurAlg} converges in all cases and achieves a score of 60.5 using uniform quantization at rank 32. {\OurAlg} stands out in its ability to consistently attain robust and improved performance by effectively preserving the starting point of pre-trained weights.

\subsection{Encoder-Decoder Model: BART}
\textbf{Models and Datasets.} We quantize BART-large model \citep{lewis-etal-2020-bart}  with {\OurAlg}, then finetune and evaluate the model on two commonly used summarization datasets: XSum \citep{Narayan2018xsum} and CNN/DailyMail\citep{hermann2015teaching}.

\noindent \textbf{Implementation Details.} We apply {\OurAlg} to weight matrices in MHA and FFN of both encoder and decoder layers. We report ROUGE 1/2/L scores, which are the metrics for summarization tasks \citep{lin-2004-rouge}. We conduct quantization experiments in both 2-bit and 4-bit scenarios. We experiment with both NormalFloat and the uniform quantization in both 2-bit and 4-bit scenarios. In each precision, we choose rank equal to 8 and 16 for a fair comparison with the full precision LoRA baseline \citep{zhang2023adaptive}. Please see Appendix \ref{sec:app_summarization} for detailed configurations.

\noindent \textbf{Main Results.} Table \ref{tab:bart_summerization_bit4} summarizes our 4-bit quantization experiment results on the XSum and CNN/DailyMail test sets. Our method consistently outperforms QLoRA at both ranks on both datasets. It even surpasses full precision LoRA at both ranks on Xsum. We will discuss this unexpected results in Section \ref{sec:discussion}. The 2-bit quantization results are shown in Table \ref{tab:bart_summerization_bit2}. Our observation is consistent with the NLU experiments, that {\OurAlg} demonstrates the convergence to reasonable results, while QLoRA does not converge. This indicates our method is robuster by narrowing the initialization gap. 

\begin{table}[htb!]
% \vspace{-2mm}
\caption{Results with 4-bit {\OurAlg} of BART-large on XSum and CNN/DailyMail. We report ROUGE-1/2/L, the higher the better. \textit{Lead-3} means choosing the first 3 sentences as the summary. {\it N.A.} indicates the model does not converge. \textit{Full FT} refers to the full fine-tuning where all parameters are tuned. We report the median over five seeds.}
\label{tab:bart_summerization_bit4}
\vspace{-4mm}
\begin{center}
\begin{small}
\begin{tabular}{c|c|c|cc}
\toprule
{\bf Quantization} & {\bf Rank} & {\bf Method} & {\bf XSum} & {\bf CNN/DailyMail} \\ 
\midrule 
\multirow{4}*{Full Precision} & \multirow{2}*{-} & {Lead-3} & {16.30/1.60/11.95} & {40.42/17.62/36.67}  \\
~  & ~ & {Full FT} & {45.14/22.27/37.25} & {44.16/21.28/40.90} \\ \cmidrule{2-5}
~ & {8} & {LoRA} & {43.40/20.20/35.20} & {44.72/21.58/41.84} \\
~ & {16} & {LoRA} & {43.95/20.72/35.68} & {45.03/21.84/42.15} \\
\midrule
\multirow{4}*{NF4} & \multirow{2}*{8} & {QLoRA} & {42.91/19.72/34.82} & {43.10/20.22/40.06} \\
~ & ~ & {\OurAlg} & {\bf 44.08/20.72/35.89} & {\bf 43.81/20.95/40.84} \\ \cmidrule{2-5}
~ & \multirow{2}*{16} & {QLoRA} & {43.29/20.05/35.15} & {43.42/20.62/40.44} \\
~ & ~ & {\OurAlg} & {\bf 44.51/21.14/36.18} & {\bf 43.96/21.06/40.96} \\
\midrule
\multirow{4}*{Uniform} & \multirow{2}*{8} & {QLoRA} & {41.84/18.71/33.74} & {N.A.} \\
~ & ~ & {\OurAlg} & {\bf 43.86/20.51/35.69} & {\bf 43.73/20.91/40.77} \\ \cmidrule{2-5}
~ & \multirow{2}*{16} & {QLoRA} & {42.45/19.36/34.38} & {43.00/20.19/40.02} \\
~ & ~ & {\OurAlg} & {\bf 44.29/20.90/36.00} & {\bf 43.87/20.99/40.92} \\
\bottomrule
\end{tabular}
\end{small}
\end{center}
\end{table}

\begin{table}[htb!]
% \vspace{-2mm}
\caption{Results with 2-bit {\OurAlg} of BART-large on XSum and CNN/DailyMail using {\bf NF2 quantization}. {\it N.A.} indicates the model does not converge. We report ROUGE-1/2/L, the higher the better. We report the median over five seeds.}
\label{tab:bart_summerization_bit2}
\vspace{-4mm}
\begin{center}
\begin{small}
\begin{tabular}{c|c|cc}
\toprule
{\bf Rank} & {\bf Method} & {\bf XSum} & {\bf CNN/DailyMail} \\ 
\midrule 
\multirow{2}*{8} & {QLoRA} & {N.A.} &  {N.A.} \\
~  & {\OurAlg} &  {39.63/16.65/31.62} &  {42.24/19.44/29.04} \\ 
\midrule
\multirow{2}*{16} & {QLoRA} &  {N.A.} &  {N.A.} \\
~  & {\OurAlg} &  {40.81/17.85/32.80} &  {42.52/19.81/39.51} \\

\bottomrule
\end{tabular}
\end{small}
\end{center}
\end{table}

\subsection{Decoder-only Model: LLAMA-2}
\textbf{Models and Datasets.} We quantize LLAMA-2-7b and LLAMA-2-13b \citep{touvron2023llama} with {\OurAlg}. We then fine-tune and evaluate the models on two NLG datasets: GSM8K \citep{cobbe2021training} and WikiText-2 \citep{merity2016pointer}. Please see Appendix~\ref{sec:app_NLG} for more details about the datasets.

\noindent \textbf{Implementation Details.} Similarly, we apply {\OurAlg} to weight matrices in MHA and FFN of all layers. In WikiText-2 evaluation, we report perplexity. In GSM8K evaluation, we extract numerical answers in the generated solutions and then calculate the accuracy using those numerical answers. We conduct experiments with both NF2 and NF4. Please see Appendix~\ref{sec:app_NLG} for detailed configurations.

\noindent \textbf{Main Results.} 
Table \ref{tab:llama} presents a summary of our experiments on LLAMA-2-7b and LLAMA-2-13b using 2-bit, 4-bit, and mixed-precision NormalFloat quantization methods on WikiText-2 and GSM8K datasets. In WikiText-2, our method consistently outperforms QLoRA across all quantization precision settings on both models. 
% For instance, when using 4-bit precision, {\OurAlg} achieves a perplexity of 5.24 for LLAMA-2-7b, surpassing even the performance of 16-bit precision LoRA. 
When dealing with the challenging 2-bit precision, where QLoRA fails to converge, {\OurAlg} manages to achieve a perplexity of 7.85. In GSM8K, our method achieves better or on par performance compared to QLoRA across different model sizes and quantization precision levels. For example, our method achieves 20.9\% accuracy using 2-bit precision, where QLoRA doesn't converge. 

We find {\OurAlg} outperforms full precision LoRA in GSM8K with LLAMA-2-13b. One possible explanation is that the lack of regularization causes overfitting on full precision LoRA fine-tuning. Therefore, we conduct full precision LoRA with weight decay on GSM8K. From Table \ref{tab:llama}, regularization helps LLAMA-2-13b full precision LoRA fine-tuning, but fails in LLAMA-2-7b. This indicates LLAMA-2-13b is prone to overfitting and quantization has implicit regularization to overcome such overfitting.

To provide a customized trade-off between the performance and precision, we also explore mixed-precision quantization where matrices in the first 4 layers are quantized using 4 bits, and the rest matrices remain 2 bits. We witness a remarkable 5.9\% accuracy boost on the GSM8K dataset using LLAMA-2-7b and a 12.7\% boost using LLAMA-2-13b. This result underscores the potential of {\OurAlg} for complex mixed-precision quantization scenarios.
% \begin{table}[htb!]
% % \vspace{-2mm}
% \caption{Results of {\OurAlg} using 4-bit and 2-bit normal float quantization of LLAMA-2 series on WikiText-2. We report perplexity (the lower the better). The rank of low-rank adapters is 64. Results with {\it N.A.} indicate the model does not converge.}
% \label{tab:llama_wikitext}
% \vspace{-4mm}
% \begin{center}
% \begin{small}
% \begin{tabular}{cccc}
% \toprule
% {\bf Method} & {\bf Bit} & {\bf LLAMA-2-7b} & {\bf LLAMA-2-13b} \\ 
% \midrule 
% {LoRA} & {16} & {5.31} &  {5.12} \\
% \midrule
% {QLoRA}   & {4} & {7.41} & {5.22} \\
% {\OurAlg} & {4} & {\bf 5.24} & {\bf 5.16} \\ 
% \midrule 
% {QLoRA}   & {2} & {N.A.} & {N.A.} \\
% {\OurAlg} & {2} & {\bf 7.85} & {\bf 7.69} \\ 
% \bottomrule
% \end{tabular}
% \end{small}
% \end{center}
% \end{table}

\begin{table}[htb!]
\vspace{-2mm}
\caption{Results of {\OurAlg} using NormalFloat for LLAMA-2 series on WikiText-2 and GSM8K. 3/2.5/2.25-bit indicates mixed-precision quantization: 4-bit precision for the first 16/8/4 layers and 2-bit precision for the rest of layers. We report the perplexity (the smaller the better) for WikiText-2 and accuracy for GSM8K. The rank of low-rank adapters is 64. {\it N.A.} indicates the model does not converge. We report the median over five random seeds.}

\vspace{-4mm}
\label{tab:llama}
\begin{center}
\begin{small}
\begin{tabular}{cc|cccc}
\toprule
\multirow{2}*{\bf Method} & \multirow{2}*{\bf Bit} & \multicolumn{2}{c}{\bf LLAMA-2-7b} & \multicolumn{2}{c}{\bf LLAMA-2-13b} \\ 
 ~ & ~ & {\bf WikiText-2}$\downarrow$ & {\bf GSM8K}$\uparrow$ & {\bf WikiText-2}$\downarrow$  & {\bf GSM8K}$\uparrow$\\
\midrule 
{LoRA} & {16} & {5.08} & {36.9} & {5.12} & {43.1} \\
{LoRA+Reg} & {16} & {--} & {34.4} & {--} & {45.3} \\
\midrule
{QLoRA}   & {4} & {5.70} & {\bf 35.1} & {5.22} & {39.9} \\
{\OurAlg} & {4} & {\bf 5.24} & {35.0} & {\bf 5.16} & {\bf 45.0} \\ 
\midrule
{QLoRA}   & {3} & {5.73} & {32.1} & {5.22} & {40.7} \\
{\OurAlg} & {3} & {\bf 5.63} & {\bf 32.9} & {\bf 5.13} & {\bf 44.4} \\ 
\midrule 
{QLoRA}   & {2.5} & {N.A.} & {N.A.} & {19.39} & {N.A.} \\
{\OurAlg} & {2.5} & {5.78} & {\bf 31.1} & {5.22} & {\bf 41.1} \\ 
\midrule 
{QLoRA}   & {2.25} & {N.A.} & {N.A.} & {N.A.} & {N.A.} \\
{\OurAlg} & {2.25} & {\bf 6.13} & {\bf 26.5} & {\bf 5.45} & {\bf 38.1} \\ 
\midrule 

{QLoRA}   & {2} & {N.A} & {N.A.} & {N.A.} & {N.A.} \\
{\OurAlg} & {2} & {\bf 7.85} & {\bf 20.9} & {\bf 7.69} & {\bf 25.4} \\ 
\bottomrule
\end{tabular}
\end{small}
\end{center}
\end{table}

% \vspace{-4mm}
% \label{tab:llama_gsmk8k}
% \begin{center}
% \begin{small}
% \begin{tabular}{cccc}
% \toprule
% {\bf Method} & {\bf Bit} & {\bf LLAMA-2-7b} & {\bf LLAMA-2-13b} \\ 
% \midrule 
% {LoRA} & {16} & {36.9} &  {36.5} \\
% \midrule
% {QLoRA}   & {4} & {\bf 35.1} & {39.9} \\
% {\OurAlg} & {4} & {35.0} & {\bf 45.0} \\ 
% \midrule 
% {QLoRA}   & {2.25} & {N.A.} & {N.A.} \\
% {\OurAlg} & {2.25} & {\bf 26.5} & {\bf 38.1} \\ 
% \midrule 
% {QLoRA}   & {2} & {N.A.} & {N.A.} \\
% {\OurAlg} & {2} & {\bf 20.9} & {\bf 25.4} \\ 
% \bottomrule
% \end{tabular}
% \end{small}
% \end{center}
% \end{table}

\subsection{Analysis}
\begin{figure}[htb]
\centering
\begin{subfigure}{0.32\columnwidth}
    \centering
    \includegraphics[width=1.0\textwidth]{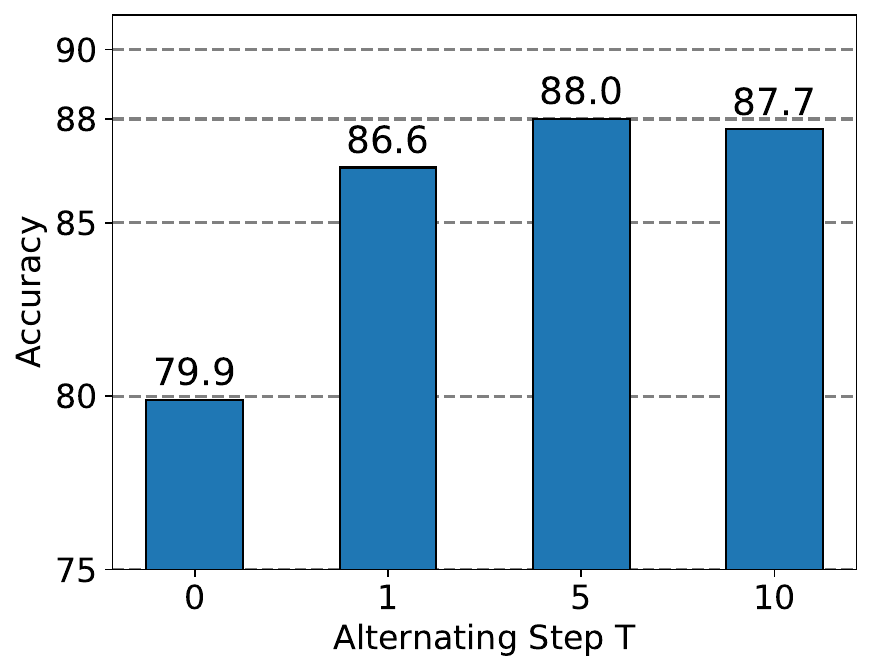}
    \vspace{-5mm}
    \caption{MNLI}
\end{subfigure}
\begin{subfigure}{0.32\columnwidth}
    \centering
    \includegraphics[width=1.0\textwidth]{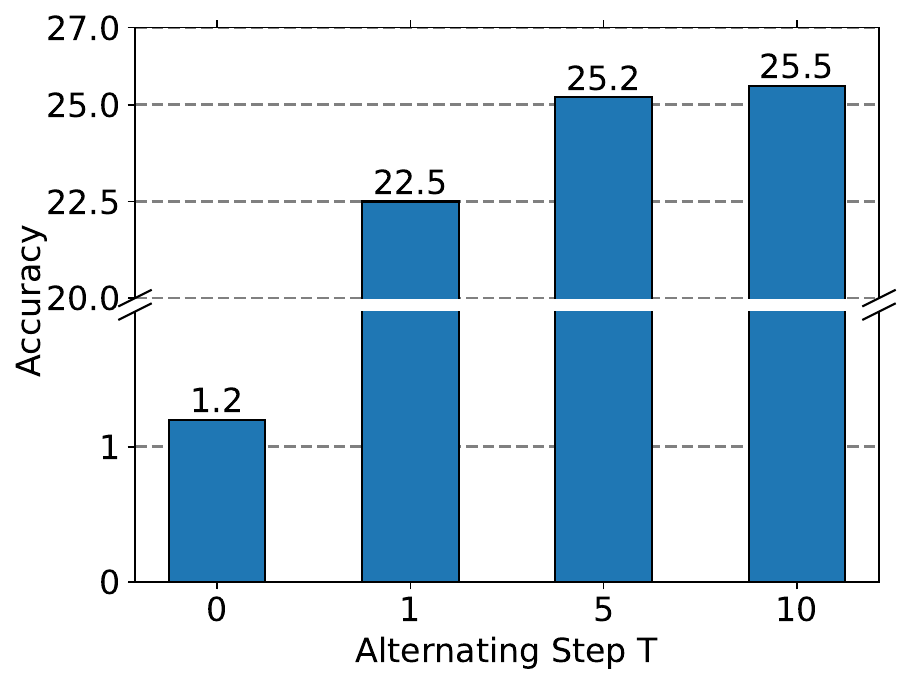}
    \vspace{-5mm}
    \caption{GSM8k}
\end{subfigure}
\begin{subfigure}{0.32\columnwidth}
    \centering
    \includegraphics[width=1.0\textwidth]{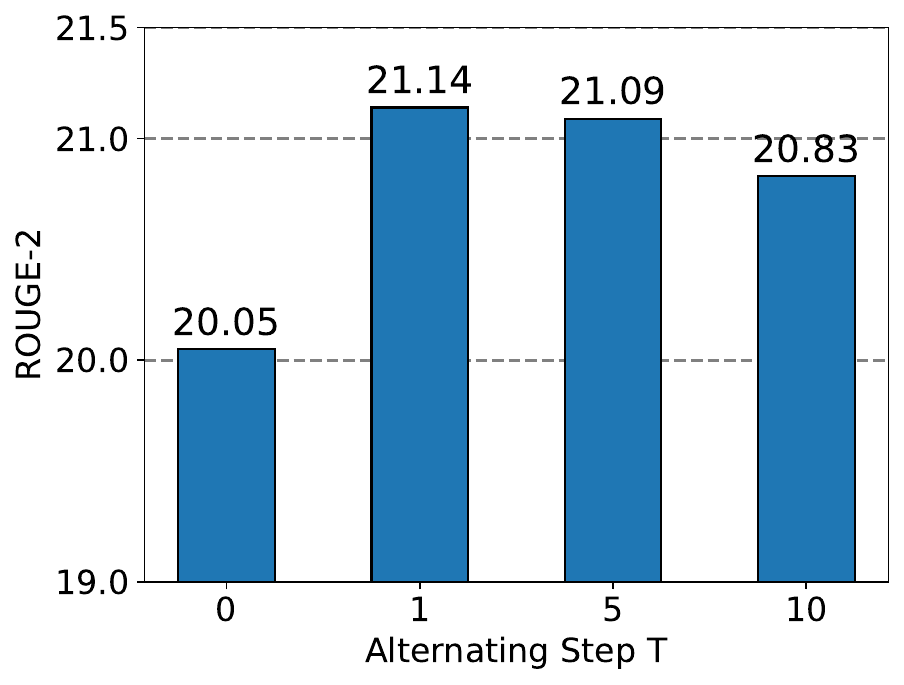}
    \vspace{-5mm}
    \caption{XSum}
\end{subfigure}
\vspace{-2mm}
\caption{
Comparison of different alternating step $T$ used in {\OurAlg}. $T=0$ indicates we use QLoRA method that initializes low-rank adapters by \eqref{eq:lora_init}. $T = 1, 5, 10 $ indicates we use different $T$ for {\OurAlg} described in Algorithm \ref{alg:alter}. {\bf Left}: Uniform 2-bit DeBERTaV3-base. {\bf Middle}: NF4 2-bit LLAMA-2-13b. {\bf Right}: NF4 BART-large.
}
\label{fig:num_iter}
\end{figure}

\textbf{Effectiveness of Alternating Optimization.}
We conduct experiments with different alternating step $T$ to verify the effectiveness of the alternating optimization and to find the best value $T$ as a hyperparameter for different models. 
Across all tasks and models, we observed that alternating optimization yields substantial improvements even with a minimal alternating step. This suggests that it rapidly narrows the discrepancy between quantized weights and pre-trained weights, making our method easy to apply. For example, our method achieves 88.0\% accuracy on MNLI-m dataset using only 5 alternating steps and 21.14 Rouge-2 score using only 1 step. Interestingly, we noticed that increasing the alternating step beyond a certain point tends to result in diminishing returns. We suspect this phenomenon occurs because, as the gap becomes smaller, it becomes more challenging for alternating optimization to consistently minimize the gap at each step. This challenge emerges because of the inherent errors introduced by the quantization method. Nevertheless, results from Figure \ref{fig:num_iter} indicate our method is not sensitive to the alternating step $T$ and is able to consistently enhance downstream fine-tuning performance.
% We remark  Figure \ref{fig:num_iter} shows that our method consistently outperforms the baseline across all tested iteration settings on different models. 
% We observed that different models achieve their best results with varying numbers of iterations, yet the performance remains stable across different numbers of iterations.
% This indicates our method is not sensitive to the number of iterations, while consistently enhancing downstream fine-tuning performance.

\section{Discussion}
\label{sec:discussion}
\textbf{Start with quantization or SVD in the alternating optimization?}
An alternative algorithm to the alternating optimization is that we first obtain the low-rank approximation $A_t, B_t$ and then obtain the quantized weight $Q_t$ by switching Line 3 and Line 4 in Algorithm \ref{alg:alter}. We note this is a valid alternative method as both still jointly minimize the objective in \eqref{eq:optimization}. Table \ref{tab:glue_alternative} summarizes the performance of this alternative method. It is noteworthy that the alternative method still outperforms QLoRA significantly, even though it is worse than the primary version. This observation underscores the potential for performance improvement by achieving a closer approximation of pre-trained weights within the low-precision regime.

% \noindent \textbf{Quantization as regularizer} We find {\OurAlg} outperforms full precision LoRA in XSum and GSM8K (see Table \ref{tab:bart_summerization_bit4} and Table \ref{tab:llama}). One possible explanation is that the lack of regularization causes overfitting on full precision LoRA fine-tuning. From Table \ref{tab:llama}, regularization helps full precision LoRA fine-tuning to overcome overfitting, but has negative effect on quantization. This implies that quantization already has regularization that prevents overfitting.
% One possible explanation for this unexpected phenomenon is that the initial low-rank adapters obtained by {\OurAlg} are non-zero while they are all zero in full precision LoRA as described in \eqref{eq:lora_init}. Such zero initialization could make the fine-tuning unstable, and therefore it performs worse than {\OurAlg}. We leave the study of the robustness of {\OurAlg} as future work.

\begin{table}[htb!]
\caption{Results of 2-bit uniformly quantized DeBERTaV3-base on part of GLUE. \OurAlg (SVD First) indicates the alternative {\OurAlg} that swiches Line 3 and Line 4 in Algorithm \ref{alg:alter}. We report the median over four random seeds. The best results on each task are shown in {\bf bold}.}
\vspace{-3mm}
\label{tab:glue_alternative}
\begin{center}
\begin{small}
\begin{tabular}{cc|cccc}
\toprule
\multirow{2}*{\bf Method} & \multirow{2}*{\bf Rank} & {\bf MNLI}  & {\bf QNLI} & {\bf SST2}   \\  
~ & ~ & {m / mm} & {Acc} & {Acc}  \\
\midrule 
{ Full FT} & - & {90.5/90.6} & {94.0} & {95.3}\\
\midrule
QLoRA & 32 & 79.9/79.5 & 83.8 & 86.6 \\
\midrule
LoftQ(SVD First) & 32 & 87.8/87.7 & 84.9 & 89.7 \\

\midrule
{\OurAlg}(Quantiztion First)& {32}  & \bf 88.0/88.1 & {\bf 92.2}&{ \bf 94.7}   \\

\bottomrule
\end{tabular}
\end{small}
\end{center}
\end{table}

% \begin{table}[htb!]
% \caption{Quantization as regularization. The results of adding weight decay to low-rank adapters of LLAMA-2-13b. Results are accuracy on GSM8K. }
% \vspace{-3mm}
% \label{tab:reg}
% \begin{center}
% \begin{small}
% \begin{tabular}{c|ccc}
% \toprule
% {\bf Quantization} & {w/ weight decay} & {w/o weight decay}\\  
% \midrule 
% {Full precision} & {45.3} & {43.1} \\
% \midrule
% {NF4} & {43.4} & {45.0} \\
% \bottomrule
% \end{tabular}
% \end{small}
% \end{center}
% \end{table}

\section{Related Work}
\textbf{Quantization-Aware Training (QAT)} is often used to obtain quantized models that are adapted in downstream tasks \citep{peri2020deploying, liu2023llm}. It involves quantization and full model fine-tuning at the same time. However, QAT requires massive training cost, such as the gradient and optimization state. Moreover, it is difficult to compute the gradient of quantized weights.
Our method, with the help of LoRA, sidesteps the aforementioned issues, providing a light approach for downstream task adaptation.

\vskip2pt

\noindent \textbf{Post-Training Quantization (PTQ)} is a category of popular quantization frameworks \citep{frantar2022gptq, xiao2023smoothquant}, which can also be used for task adaptation. It calibrates the high-precision model with a small subset of the training dataset. Therefore, the subsequent quantization is guided by the training dataset, providing task-specific quantized models. Besides, it does not involve any gradient backpropagation, so it is cost-efficient. However, it usually results in lower accuracy compared to QAT. 

\section{Conclusion}
We propose {\OurAlg}, a quantization framework for LLMs, which alternatively applies quantization and low-rank approximation to the original high-precision pre-trained weights, to obtain an initialization for the subsequent LoRA fine-tuning. Experiments on natural language understanding, question answering, summarization, and natural language generation show that our framework remarkably surpasses existing methods, e.g., QLoRA, for quantizing encoder-only, encoder-decoder, and decoder-only models. We have not observed our method exhibiting worse performance over QLoRA. Moreover, our quantization framework demonstrates effectiveness and robustness particularly in low-bit quantization regimes, e.g., the 2-bit level.

\bibliography{main}
\bibliographystyle{ims}

% %%%%%%%%%%%%%%%%%%%%%%%%%%%%%%%%%%%%%%%%%%%%%%%%%%%%%%%%%%%%%%%%%%%%%%%%%%%%%%%
% %%%%%%%%%%%%%%%%%%%%%%%%%%%%%%%%%%%%%%%%%%%%%%%%%%%%%%%%%%%%%%%%%%%%%%%%%%%%%%%
% % APPENDIX
% %%%%%%%%%%%%%%%%%%%%%%%%%%%%%%%%%%%%%%%%%%%%%%%%%%%%%%%%%%%%%%%%%%%%%%%%%%%%%%%
% %%%%%%%%%%%%%%%%%%%%%%%%%%%%%%%%%%%%%%%%%%%%%%%%%%%%%%%%%%%%%%%%%%%%%%%%%%%%%%
\newpage
\appendix
\section{Model Compression Ratio and Memory Footprint}
\label{app:model_compression_ratio}

We report the compression ratio after applying {\OurAlg} in Table \ref{tab:model-arch1}. It is defined as 
\begin{align*}
    \text{compression ration} = \frac{\text{backbone size}+\text{LoRA adapter size}}{\text{pre-trained size}}.
\end{align*}

We also measure the GPU memory cost during training. Given that GPU memory varies by models, tasks, sequence lengths, batch sizes, etc. We report LLAMA-2 on GSM8K as an example in Table \ref{tab:mem_footprint}.

\begin{table}[htb]
\centering
\caption{Compression ratios of backbones.}
\vspace{-0.1in}
\begin{tabular}{c|c|c|c|c|c}
\toprule 
\multirow{2}*{\bf Model} & {\bf Compression} & {\bf Trainable}  & \multirow{2}*{\bf Rank} & \multirow{2}*{\bf Bits} & {\bf Quantization}     \\
~ & {\bf ratio (\%)} & {\bf ratio (\%)} & ~ & ~ & {\bf method} \\
\midrule
DeBERTaV3-base & 15.6 & 3.1 & 16 & 2& Uniform \\
DeBERTaV3-base & 18.8 & 6.3 & 32 & 2& Uniform \\
DeBERTaV3-base & 17.2 & 3.1 & 16 & 2& NF2 \\
DeBERTaV3-base & 20.4 & 6.3 & 32 & 2& NF2 \\
BART-large &  15.3   & 1.2 & 8 & 4  &     NF2      \\
BART-large &  16.7   & 2.5 & 16 & 4  &     NF2      \\
BART-large &  27.8   & 1.2 & 8 & 4  &     NF4      \\
BART-large &  29.0   & 2.5 & 16 & 4  &     NF4      \\
BART-large &  26.2  & 1.2 & 8 & 4  &    Uniform      \\
BART-large &  27.5  & 2.5 & 16 & 4  &     Uniform      \\
LLAMA-2-7b & 16.6 & 2.4 &64 & 2& Nf2\\
LLAMA-2-7b & 29.0 & 2.4 & 64& 4& Nf4\\
LLAMA-2-13b & 16.0 &1.9 &64 & 2& Nf2\\
LLAMA-2-13b & 28.5 &1.9 & 64& 4& Nf4\\
\bottomrule
\end{tabular}
\label{tab:model-arch1}
\end{table}

\begin{table}[htb]
\centering
\caption{GPU memory footprint}
\begin{tabular}{c|c|c|c|c}
\toprule
{\bf Model} & {\bf Dataset}  & {\bf Seq length} & {\bf Batch size}  & {\bf GPU Mem}   \\
\midrule
LLAMA-2-7b & GSM8K & 384 & 1 & 15GB \\
LLAMA-2-13b & GSM8K & 384 & 1 & 24GB \\
\bottomrule
\end{tabular}
\label{tab:mem_footprint}
\end{table}

\section{Quantization Time}
We report the execution time of {\OurAlg} applying to a single weight matrix in Table \ref{tab:model-arch2}. The time is tested on Intel(R) Xeon(R) CPU E5-2650 v4 @ 2.20GHz.
\begin{table}[htb]
\centering
\caption{Execution time of {\OurAlg} applying to different weight matrices.}
\begin{tabular}{c|c|c|c|c}
\toprule
{\bf Model} & {\bf Size} & {\bf Step $T$} & {\bf Quantization method}  & {\bf Time}   \\
\midrule
DeBERTaV3-base & $768\times 768$ & 5& Uniform & 1s\\
BART-large & $1024\times 1024$ & 5 & NF4 & 1s\\
LLAMA-2-7b & $4096\times 4096$ & 5 & NF4 & 21s\\
LLAMA-2-13b & $5120\times 5120$ & 5 & NF4 & 43s\\
\bottomrule
\end{tabular}
\label{tab:model-arch2}
\end{table}

\section{GLUE Dataset Statistics}
\label{app:dataset-glue}

We present the dataset statistics of GLUE \cite{wang2018glue} in the following table. 
\begin{table*}[htb]
	\begin{center}
		\begin{tabular}{l|l|c|c|c|c|c}
			\toprule 
			\bf Corpus & \bf Task &\bf \#Train &\bf \#Dev &\bf \#Test &\bf \#Label &\bf Metrics\\ \midrule
			\multicolumn{6}{@{\hskip1pt}r@{\hskip1pt}}{Single-Sentence Classification (GLUE)} \\ \hline
			CoLA & Acceptability&8.5k & 1k & 1k & 2 & Matthews corr\\ \hline
			SST & Sentiment&67k & 872 & 1.8k & 2 & Accuracy\\ \midrule
			\multicolumn{6}{@{\hskip1pt}r@{\hskip1pt}}{Pairwise Text Classification (GLUE)} \\ \hline
			MNLI & NLI& 393k& 20k & 20k& 3 & Accuracy\\ \hline
			RTE & NLI &2.5k & 276 & 3k & 2 & Accuracy \\ \hline
			% WNLI & NLI &634& 71& 146& 2 & Accuracy \\ \hline
			QQP & Paraphrase&364k & 40k & 391k& 2 & Accuracy/F1\\ \hline
			MRPC & Paraphrase &3.7k & 408 & 1.7k& 2&Accuracy/F1\\ \hline
			QNLI & QA/NLI& 108k &5.7k&5.7k&2& Accuracy\\ \midrule
			\multicolumn{5}{@{\hskip1pt}r@{\hskip1pt}}{Text Similarity (GLUE)} \\ \hline
			STS-B & Similarity &7k &1.5k& 1.4k &1 & Pearson/Spearman corr\\ \bottomrule
			%			\multicolumn{6}{@{\hskip1pt}r@{\hskip1pt}}{Pairwise Text Classification} \bottomrule %\\ \hline
			% 			SNLI & NLI& 549k &9.8k&9.8k&3& Accuracy\\ \hline
			% 			SciTail & NLI& 23.5k &1.3k&2.1k&2& Accuracy\\ \hline
			% 			ANLI & NLI& 163k &3.2k&3.2k&3& Accuracy\\ \hline
		\end{tabular}
	\end{center}
	\vskip -0.05in
	\caption{Summary of the GLUE benchmark.}
	\label{tab:glue}
\end{table*}

GLUE includes two single-sentence classification tasks: SST-2 \citep{socher-etal-2013-recursive} and CoLA \citep{warstadt-etal-2019-neural}, and three similarity and paraphrase tasks: MRPC \citep{dolan-brockett-2005-automatically}, STS-B \citep{cer-etal-2017-semeval}, and QQP. GLUE also includes four natural language inference tasks in GLUE: MNLI \citep{williams-etal-2018-broad}, QNLI \citep{rajpurkar-etal-2016-squad}, RTE \citep{Dagan2007ThePR,BarHaim2006TheSP,giampiccolo-etal-2007-third,bentivogli2009fifth}, and WNLI \citep{levesque2012winograd}.

\section{Natural Language Understanding}
\label{sec:app_nlu}
\subsection{GLUE with 4-bit}
We show the 4-bits results in the Table \ref{tab:glue_nf4}. Both methods can achieve performance close to full-finetuning.
\label{sec: app_nlu_nf4}
\begin{table}[htb]
\vspace{-3mm}
\caption{Results with 4-bit {\OurAlg} of DeBERTaV3-base models on GLUE development set using NF4 quantization. We report the median over four seeds. Results with N.A. indicate the model does not converge. The best results on each dataset are shown in bold}
\vspace{1mm}
\label{tab:glue_nf4}
\begin{center}
\begin{tabular}{c|c|cccc}
\toprule
\multirow{2}*{\bf Method} & \multirow{2}*{\bf Rank} & {\bf MNLI} & {\bf SST-2} & {\bf QNLI} & {\bf ANLI}   \\  
~ & ~ & {m / mm} & {Acc} & {Acc} & {Acc} \\
\midrule 
{ Full FT} & - & {90.5/90.6} & {95.3} & {94.0} & {59.8}\\
\midrule
QLoRA & 32 & 89.9/89.9 & \bf 95.3 & \bf 94.2 & {59.4}\\

\midrule
{\OurAlg}& {32}  & \bf 89.9/90.0 & {\bf 95.3}&{ 94.1}  & {\bf59.9}  \\

\bottomrule
\end{tabular}
\end{center}

\end{table}

\subsection{Training Details} 
\label{sec:app_nlu_details}
{\bf Implementation Details.} The implementation of {\OurAlg} is based on publicly available Huggingface \citep{NEURIPS2019_9015} code-base \footnote{\url{https://github.com/huggingface/transformers/tree/main/examples/pytorch}}. 

\noindent {\bf Hyper-parameter Details.} 
We select the learning rate of $\{1\times10^{-5}, 5\times10^{-5}, 1\times10^{-4}, 5\times10^{-4}\}$, and use the selected learning rate for both uniform quantization experiments and nf2 quantization experiments. We use batch size of 32 for all GLUE tasks and ANLI. We use batch size of 16 for SQuADv1.1. We use {\OurAlg} of 5 iterations for all GLUE tasks. 

Table \ref{tab:app_glue_setup_uniform} summarizes the detailed hyperparameters for each task used in training DeBERTaV3-base using uniform quantization. Table \ref{tab:app_glue_setup_nf2}
 summarizes the detailed hyperparameters for each task used in training DeBERTaV3-base using nf2 quantization. 

\begin{table*}[htb]
\vspace{-1mm}
\caption{Hyper-parameter setup of {\OurAlg} for GLUE benchmark for training DeBERTaV3-base using Uniform quantization.}
%\vspace{-1mm}
\label{tab:app_glue_setup_uniform}
\begin{center}
\scalebox{0.68}{
\begin{small}
\begin{tabular}{c|cccccccccc}
\toprule
 {\bf Hyper-parameter} & {\bf MNLI} & {\bf RTE} & {\bf QNLI}  & {\bf MRPC} & {\bf QQP } & {\bf SST-2} & {\bf CoLA} & {\bf STS-B} & {\bf SQuADv1.1}&{\bf ANLI} \\ 
\midrule 
 \# epochs &   5  &  20 & 10 & 60 & 10 & 10 & 60 & 60&10 &12 \\
Learning rate & $ 1\times 10^{-4}$ & $5\times 10^{-4}$ & $ 5\times 10^{-5} $ & $ 1\times 10^{-4} $  & $ 5\times 10^{-5} $ & $ 5\times 10^{-5}$ & $ 5\times 10^{-5} $ & $5\times 10^{-5}$ & $5\times 10^{-5}$ & $5\times 10^{-5}$ \\

\bottomrule
\end{tabular}
\end{small}}
\end{center}
%\vspace{-1mm}
\end{table*}

\begin{table*}[htb]
\vspace{-1mm}
\caption{Hyper-parameter setup of {\OurAlg} for GLUE benchmark for training DeBERTaV3-base using NF2 quantization.}
%\vspace{-1mm}
\label{tab:app_glue_setup_nf2}
\begin{center}
\scalebox{0.68}{
\begin{small}
\begin{tabular}{c|cccccccccc}
\toprule
 {\bf Hyper-parameter} & {\bf MNLI} & {\bf RTE} & {\bf QNLI}  & {\bf MRPC} & {\bf QQP } & {\bf SST-2} & {\bf CoLA} & {\bf STS-B} & {\bf SQuADv1.1}&{\bf ANLI} \\ 
\midrule 
 \# epochs &   5  &  20 & 10 & 60 & 10 & 10 & 60 & 60&10 &12 \\
Learning rate & $ 1\times 10^{-4}$ & $5\times 10^{-5}$ & $ 5\times 10^{-5} $ & $ 1\times 10^{-4} $  & $ 5\times 10^{-5} $ & $ 5\times 10^{-5}$ & $ 5\times 10^{-5} $ & $1\times 10^{-4}$ & $5\times 10^{-5}$ & $5\times 10^{-5}$ \\

\bottomrule
\end{tabular}
\end{small}}
\end{center}
%\vspace{-1mm}
\end{table*}
% % \section{Question Answering}
% % \label{sec:app_squad}

% % \subsection{Dataset}
% % Following \citet{sanh2020movement}, we also choose SQuAD v1.1 \cite{rajpurkar2016squad} to evaluate the performance of {\OurAlg} on question answering task.  

% % \subsection{Training Details}
% % %\vspace{0.05in} \noindent
% % %{\bf Implementation Details.} 
% % We set the batch size as 16, the number of epochs for fine-tuning as 10, the optimizer as AdamW and the learning rate as $ 5\times 10^{-5} $ for all experiments. Similarly, we follow the pruning schedule of \citet{zhang2022platon},i.e. we take the same initial warm up steps and final warm up steps. We use the same settings for all sparsities. The hyperparameters are summarized specifically in Table \ref{tab:app_squad_setup}. We use the hyperparameters in Table \ref{tab:app_squad_setup} for pruning both DeBERTaV3-base and BERT-base.
% % % Please see Appendix~\ref{sec:app_squad} for more details about hyperparameter setup.  

% % \begin{table*}[htb!]
% % \vspace{-2mm}
% % \caption{Hyper-parameter setup of {\OurAlg} on question answering tasks (SQuAD v1.1, \citet{rajpurkar2016squad}).}
% % \vspace{3mm}
% % \label{tab:app_squad_setup}
% % \begin{center}
% % \begin{small}
% % \begin{tabular}{l|cccccccc}
% % \toprule
% % \multirow{1}*{\bf Task} & \# epochs & Batch size  & Learning rate & $ t_i $ & $ t_f $ & $r$ & $\beta$ \\ 
% % \midrule 
% % SQuAD  & 10&16& $ 5\times 10^{-5} $ &5400&22000&5\% & 0.85 \\
% % \bottomrule
% % \end{tabular}
% % \end{small}
% % \end{center}
% % %\vspace{-1mm}
% % \end{table*}

\section{Summarization}
\label{sec:app_summarization}
\subsection{Training Details}
We choose Adam as the optimizer and try learning rate from\{$1\times10^{-5}, 5\times10^{-5}, 7\times10^{-5}, 2\times10^{-4}, 3\times10^{-4}, 4\times10^{-4}\}$. We show the optimal learning rate for different settings in Table \ref{tab:app_bart_setup_cnn}.  We use {\OurAlg} of 1 iteration for all BART-large experiments. Table \ref{tab:app_bart_setup_cnn} and Table \ref{tab:app_bart_setup_xsum} summarize the learning rate and other hyper-parameters for CNN/DailyMail and XSum.

\begin{table}[htb]
\caption{Hyper-parameter setup of {\OurAlg} BART-large on  CNN/DailyMail}
\vspace{-5mm}
\label{tab:app_bart_setup_cnn}
\begin{center}

\begin{tabular}{c|cc|cc|cc}
\toprule
\multirow{2}*{\bf Hyperparameter} & \multicolumn{2}{c}{\bf NF4} & \multicolumn{2}{c}{\bf 4-bit Uniform}& \multicolumn{2}{c}{\bf NF2} \\ 

~ & rank8& rank16& rank8&rank16 &rank8&rank16 \\
\midrule 
% \cmidrule{2-7}
Learning rate&2e-4&2e-4&2e-4&3e-4 &2e-4&2e-4  \\
\midrule 
Epoch &15&15&15&15 &15&15  \\
\midrule 
Batch size &64&64&64&64&64&64  \\
\bottomrule
\end{tabular}

\end{center}
%\vspace{-1mm}
\end{table}

\begin{table*}[h!]
\caption{Hyper-parameter setup of {\OurAlg} BART-large on XSum}
\vspace{-5mm}
\label{tab:app_bart_setup_xsum}
\begin{center}
\begin{tabular}{c|cc|cc|cc}
\toprule
\multirow{2}*{\bf Hyperparameter} & \multicolumn{2}{c}{\bf NF4} & \multicolumn{2}{c}{\bf 4-bit Uniform}& 
\multicolumn{2}{c}{\bf NF2} \\ 
 ~ & rank8& rank16& rank8&rank16 &rank8&rank16 \\
\midrule 
Learning rate & {2e-4} & {2e-4} & {2e-4} & {2e-4} & {2e-4} & {2e-4}  \\
\midrule 
Epoch &25&25&25&25 &25&25  \\
\midrule 
Batch size &32&32&32&32&32&32  \\
\bottomrule
\end{tabular}

\end{center}
%\vspace{-1mm}
\end{table*}

\section{Natural Language Generation}
\label{sec:app_NLG}
We set the batch size as 32 for WikiText-2 and 16 for GSM8K. We train 2 epochs on WikiText-2 and 6 epochs on GSM8K. We select learning rate from\{$1\times10^{-5}, 5\times10^{-5}, 7\times10^{-5}, 1\times10^{-4}, , 3\times10^{-4}, 4\times10^{-4}\}$.
Specific settings are summarized in Table \ref{tab:app_llama_gsm8k} and Table \ref{tab:app_llama_wikitex}.

\begin{table}[htb]
\caption{Hyper-parameter setup of {\OurAlg} LLAMA-2-series on GSM8K}
\vspace{-5mm}
\label{tab:app_llama_gsm8k}
\begin{center}
\begin{tabular}{c|c|ccc}
\toprule
{\bf Model}& {\bf Hyperparameter}&  {\bf NF4} & {\bf NF2}& {\bf Mixed-precision}  \\ 
\midrule 
LLAMA-2-7b&learning rate&$3\times 10^{-4}$&$3\times 10^{-4}$&$3\times 10^{-4}$ \\
\midrule
LLAMA-2-13b&learning rate&$1\times 10^{-4}$ & $1\times 10^{-4}$&$3\times 10^{-4}$ \\
\bottomrule
\end{tabular}
\end{center}
%\vspace{-1mm}
\end{table}

\begin{table}[htb]
\caption{Hyper-parameter setup of {\OurAlg} LLAMA-2-series on WikiText-2}
\vspace{-5mm}
\label{tab:app_llama_wikitex}
\begin{center}
\begin{tabular}{c|c|ccc}
\toprule
\bf Model& {\bf Hyperparameter}&  {\bf NF4} & {\bf NF2}& {\bf Mixed-precision}  \\ 
\midrule 
LLAMA-2-7b&learning rate&$3\times 10^{-4}$&$3\times 10^{-4}$&$3\times 10^{-4}$ \\
\midrule
LLAMA-2-13b&learning rate&$1\times 10^{-4}$&$1\times 10^{-4}$&$3\times 10^{-4}$ \\
\bottomrule
\end{tabular}
\end{center}
%\vspace{-1mm}
\end{table}

\section{Comparison to Pruning}
\label{sec:app_compare_pruning}
Pruning is also a widely used compression method. Here we compare {\OurAlg} with the state-of-the-art pruning method \citet{li2023losparse}. We show the comparison in Table \ref{tab:glue_compare_prunning}. We can see our method significantly outperforms the pruning methods on DeBERTaV3-base model. We also remark that {\OurAlg} can consistently reduce the memory of both training and storage. In contrast, pruning requires training the entire full-precision matrix, which implies that it can not achieve any memory savings during the training stage.
\begin{table}[htb!]
\vspace{-3mm}
\caption{Results of {\OurAlg} using 2-bits uniform quantization compared with LoSparse with DeBERTaV3-base models on some of GLUE development sets. Here {\it Ratio} is the proportion of total remaining weights. Results with {\it N.A.} indicate the model does not converge.}
\vspace{1mm}
\label{tab:glue_compare_prunning}
\begin{center}
\begin{small}
\begin{tabular}{l|l|ccc}
\toprule
\multirow{2}*{\bf Method} & \multirow{2}*{\bf Ratio} & {\bf MNLI} & {\bf SST-2} & {\bf QNLI}   \\  
~ & ~ & {m / mm} & {Acc} & {Acc} \\
\midrule 
{ Full FT} & {100\%} & {90.5 / 90.6} & {95.3} & {94.0}\\
\midrule
\multirow{2}*{LoSparse} & {15\%} &{83.3/82.9} & 87.6 & 90.4\\
~ & {20\%}& {84.5/83.8 } & {91.7} & {88.6} \\
\midrule
\multirow{2}*{{\OurAlg}} & {15.6\%}  &{\bf 87.3/87.1} & {\bf 94.0}&{\bf 90.6}    \\
~ & {18.8\%} & {\bf88.0/88.1} &{\bf94.7}   &  {\bf92.4} \\
\bottomrule
\end{tabular}
\end{small}
\end{center}

\end{table}

\section{Extension to Convolutional Layers}
Low-rank adapters can also be applied to convolutional layers. Given an input feature map $X \in \RR^{h \times w \times c_1}$ and $c_2$ 2D convolutional kernels $K_i \in \RR^{c_1 \times d \times d}, i=1, 2, ..., c_2$, the output of the convolutional layer is
\begin{align}
    Y = \mathrm{stack}(X \otimes K_1,..., X \otimes K_{c_2}),
    \label{eq:conv}
\end{align}
where $Y \in \RR^{h \times w \times c_2}$ and $\otimes$ denotes the 2D convolution operation.

We can reformulate Equation \eqref{eq:conv} into matrix multiplication as
\begin{align*}
    Y = Z \times H^{\top},
\end{align*}
where $Z \in \RR^{hw \times c_1 d^2}, H \in \RR^{c_2 \times c_1 d^2}$, by extending and flattening the input $X$ together with concatenating and flattening kernels. 
We first extend a vector $x_{i, j} \in \RR^{c_1}$ by its neighbor vectors within the kernel window: 
\begin{align*}
    x_{i, j}^{'} = \rm{Concat}(x_{i-\frac{d}{2}, j-\frac{d}{2}}, ..., x_{i+\frac{d}{2}, j+\frac{d}{2}}).
\end{align*}
Now, $X$ becomes $X' \in \RR^{h \times w \times c_1 d^2}$. We then flatten $X'$ into $Z \in \RR^{hw \times c_1 d^2}$.
For kernels, we first concatenate $\{K_1, ..., K_{c_2}\}$ into $H' \in \RR^{c_2 \times c_1 \times d \times d}$. We then flatten $H'$ into $H$.

Note that $H$ can be approximated by a low-rank matrix
\begin{align*}
    R = UV^{\top},
\end{align*}
where $U \in \RR^{c_2 \times r}, V \in \RR^{c_1 d^2 \times r}, r \ll \min\{c_2, c_1 d^2\}$ by SVD. Therefore, the original convolution layer can be approximated as 
\begin{align}
    \hat{Y} &= Z \times (UV^{\top})^{\top} \\
              &= (Z \times V) \times U^{\top} \\
              &= M \times U^{\top}.
\end{align}

Note that $Z \times V$ can be restored into a convolution operation where we have $r$ kernels $D_i \in \RR^{c_1 \times d \times d}, i=1, 2,, ..., r$ and $M \times U^{\top}$ can also be restored into a convolution operation where we have $c_2$ kernels $U_i \in \RR^{r \times 1 \times 1}, i=1, 2,, ..., c_2$.

\end{document}